\documentclass[10pt,twocolumn,twoside]{IEEEtran}
\usepackage{amsmath,epsfig}
\usepackage{amssymb}
\usepackage{graphicx}
\usepackage{epstopdf}
\usepackage{url}
\usepackage[noadjust]{cite}

\begin{document}

\title{ Preconditioned Stochastic Gradient Descent}
\author{ Xi-Lin Li   
\thanks{Independent researcher, 1521 Sierraville Ave, San Jose, CA 95132 (email: lixilinx@gmail.com). }	
	}

\maketitle

\begin{abstract}
Stochastic gradient descent (SGD) still is the workhorse for many practical problems. However, it converges slow, and can be difficult to tune. It is possible to precondition SGD to accelerate its convergence remarkably. But many attempts in this direction either aim at solving specialized problems, or result in significantly more complicated methods than SGD. This paper proposes a new method to adaptively estimate a preconditioner such that the amplitudes of perturbations of preconditioned stochastic gradient match that of the perturbations of parameters to be optimized in a way comparable to Newton method for deterministic optimization. Unlike the preconditioners based on secant equation fitting as done in deterministic quasi-Newton methods, which assume positive definite Hessian and approximate its inverse, the new preconditioner works equally well for both convex and non-convex optimizations with exact or noisy gradients. When stochastic gradient is used, it can naturally damp the gradient noise to stabilize SGD. Efficient preconditioner estimation methods are developed, and with reasonable simplifications, they are applicable to large-scale problems. Experimental results demonstrate that equipped with the new preconditioner, without any tuning effort, preconditioned SGD can efficiently solve many challenging problems like the training of a deep neural network or a recurrent neural network requiring extremely long term memories.
\end{abstract}


\section{Introduction}

Stochastic gradient descent (SGD) has a long history in signal processing and machine learning \cite{Godard80,Rumelhart86,Widrow85,bptt,LeCun98,Cardoso96, Amari96}. In adaptive signal processing, an exact gradient might be unavailable in a time-varying setting, and typically it is replaced with instantaneous gradient, a stochastic gradient with mini-batch size 1 \cite{Godard80,Widrow85}. In machine learning like the training of neural networks, deterministic gradient descent is either expensive when the training data are large, or unnecessary when the training samples are redundant \cite{LeCun98}. SGD keeps to be a popular choice due to its simplicity and proved efficiency in solving large-scale problems. However, SGD may converge slow and is difficult to tune, especially for large-scale problems. When the Hessian matrix is available and small, second order optimization methods like the Newton method might be the best choice, but in practice, this is seldom the case. For example, calculation of Hessian for a fairly standard feedforward neural network can be much more complicated than its gradient evaluation \cite{Wray91}. In deterministic optimization, quasi-Newton methods and (nonlinear) conjugate gradient methods are among the most popular choices, and they converge fast once the solution is located in a basin of attraction. However, these methods require a line search step, which can be problematic when the cost function cannot be efficiently evaluated, a typical scenario where SGD is used. Still, they are applied with successes to machine learning problems like neural network training, and are made available in standard toolboxes, e.g., the Matlab neural network toolbox \cite{matlab}. The highly specialized Hessian-free neural network training methods in \cite{Martens2012_hessian_free} represent the state of the art in this direction. There are attempts to adapt the deterministic quasi-Newton methods to stochastic optimization \cite{stochastic_newton1,convex_quasi_newton,stochastic_newton2}. However, due to the existence of gradient noise and infeasibility of line search in stochastic optimization, the resultant methods either impose strong restrictions such as convexity on the target problems, or are significantly more complicated than SGD. On the other hand, numerous specialized SGD methods are developed for different applications. In blind source separation and independent component analysis, relative (natural) gradient is proposed to replace the regular gradient in SGD \cite{Cardoso96, Amari96}. However, in general the natural gradient descent using metrics like Fisher information can be as problematic as Newton method for large-scale problems. In neural network training, a number of specialized methods are developed to improve the convergence of SGD, and to name a few, the classic momentum method and Nesterov's accelerated gradient, the RMSProp method and its variations, various step size control strategies, pre-training, clever initialization, and a few recent methods coming up with element-wise learning rates \cite{rmsprop, Sutskever2013, no_more_pesky_mu, Equilibrated_mu,Chunyuan,Carlson}. Clearly, we need a stochastic optimization method that is as simple and widely applicable as SGD, converges as fast as a second order method, and lastly but not the least, is user friendly, requiring little tuning effort.  

In this paper, we carefully examine SGD in the general non-convex optimization setting. We show that it is possible to design a preconditioner that works well for both convex and non-convex problems, and such a preconditioner can be estimated exclusively from the noisy gradient information. However, when the problem is non-convex, the preconditioner cannot be estimated following the conventional ways of Hessian estimation as done in the quasi-Newton methods. For the general non-convex optimization, a new method is required to estimate the  preconditioner, which is not necessarily the inverse of Hessian, but related to it. This new preconditioner has several desired properties. It reduces the eigenvalue spread of preconditioned SGD to speed up convergence, scales the stochastic gradient in a way comparable to Newton method such that step size selection is trivial, and lastly, it has a built-in gradient noise suppression mechanism to stabilize preconditioned SGD when the gradient is heavily noisy. Practical implementation methods are developed, and applications to a number of interesting problems demonstrate the usefulness of our methods.            

\section{Background}

\subsection{SGD}

Although this paper focuses on stochastic optimization, deterministic gradient descent can be viewed as a special case of SGD without gradient noise, and the theories and methods developed in this paper are applicable to it as well. Here, we consider the minimization of cost function
\begin{equation}
	f(\pmb \theta) = E[\ell(\pmb \theta, \pmb z)],
\end{equation}
where $\pmb \theta$ is a parameter vector to be optimized, $\pmb z$ is a random vector,  $\ell$ is a loss function, and $E$ takes expectation over $\pmb z$. For example, in the problem of classification using neural network, $\pmb \theta$ represents the vector containing all the tunable weights in the neural network, $\pmb z=(\pmb x, y)$ is the pair of feature vector $\pmb x$ and class label $y$, and $\ell$ typically is a differentiable loss like the mean squared error, or the cross entropy loss, or the (multi-class) hinge loss. Gradient descent can be used to learn $\pmb \theta$ as
\begin{equation}
	\pmb \theta^{[\rm new]} = \pmb \theta^{[\rm old]} - \mu \pmb g(\pmb \theta^{[\rm old]}),
\end{equation}
where $\mu>0$ is a positive step size, and 
\begin{equation}\label{expect_g}
	\pmb g(\pmb \theta) =  E\left[ \frac{\partial \ell(\pmb \theta, \pmb z) }{\partial \pmb \theta}\right]
\end{equation}
is the gradient of $f(\pmb \theta)$ with respect to $\pmb \theta$. Evaluation of the expectation in (\ref{expect_g}) may be undesirable or not possible. In SGD, this expectation is approximated with sample average, and thus leading to the following update rule for $\pmb \theta$,
\begin{eqnarray}\label{sgd}
	\hat{\pmb g}(\pmb \theta) & = & \frac{1}{n}\sum_{i=1}^n  \frac{\partial \ell(\pmb \theta, \pmb z_i) }{\partial \pmb \theta} , \\ \label{sdg_theta}
	\pmb \theta^{[\rm new]} & = & \pmb \theta^{[\rm old]} - \mu \hat{\pmb g}(\pmb \theta^{[\rm old]}) ,
\end{eqnarray}
where the hat $\wedge$ suggests that the variable under it is estimated, $n\ge 1$ is the mini-batch size, and $\pmb z_i$ denotes the $i$th sample, typically randomly drawn from the training data. Now $\hat{\pmb g}(\pmb \theta)$ is a random vector, and it is useful to rewrite it as
\begin{equation}\label{error_1}
	\hat{\pmb g}(\pmb \theta) = \pmb g(\pmb \theta) + \pmb \epsilon'
\end{equation}
to clearly show its deterministic and random parts, where random vector $\pmb \epsilon' $ models the approximation error due to replacing expectation with sample average. Although being popular due to its simplicity, SGD may converge slow and the selection of $\mu$ is nontrivial, as revealed in the following subsection.       

\subsection{Second Order Approximation}

We consider the following second order approximation of $f(\pmb \theta)$ around a point $\pmb \theta_0$,
\begin{equation}\label{soa}
	f(\pmb \theta) \approx f(\pmb \theta_0) + \pmb g_0^T (\pmb \theta - \pmb \theta_0)  + \frac{1}{2} (\pmb \theta - \pmb \theta_0) ^T \pmb H_0 (\pmb \theta - \pmb \theta_0)  ,
\end{equation}
where superscript $T$ denotes transpose, $\pmb g_0 = \pmb g(\pmb \theta_0)$ is the gradient at $\pmb\theta_0$, and 
$$\pmb H_0 = \left. \frac{\partial^2 f(\pmb \theta)}{\partial \pmb \theta ^T \partial \pmb \theta }\right|_{\pmb \theta = \pmb \theta_0}$$ 
is the Hessian matrix at $\pmb \theta_0$. Note that $\pmb H_0$ is symmetric by its definition. With this approximation, the gradients of $f(\pmb \theta)$ with respect to $\pmb \theta$ around $\pmb \theta_0$ can be evaluated as
\begin{eqnarray}\label{error_2}
	\pmb g(\pmb \theta) & \approx & \pmb g_0 + \pmb H_0 (\pmb \theta - \pmb \theta_0), \\ \label{new_hat_g}
	\hat{\pmb g}(\pmb \theta) & = & \pmb g_0 + \pmb H_0 (\pmb \theta - \pmb \theta_0) + \pmb \epsilon,
\end{eqnarray}
where $\pmb \epsilon$ contains the errors introduced in both (\ref{error_1}) and (\ref{error_2}). Using (\ref{new_hat_g}), around $\pmb \theta_0$, the learning rule (\ref{sdg_theta}) turns into the following linear system,
\begin{equation}\label{iteration_no_P}
	\pmb \theta^{[\rm new]} = (\pmb I - \mu \pmb H_0) \pmb \theta^{[\rm old]} - \mu(\pmb g_0 - \pmb H_0 \pmb \theta_0 + \pmb \epsilon),
\end{equation}
where $\pmb I$ is a conformable identity matrix. Behaviors of such a linear system largely depend on the selection of $\mu$ and the distribution of the eigenvalues of $\pmb I - \mu \pmb H_0$. In practice, this linear system may have a large dimension and be ill-conditioned. Furthermore, little is known about $\pmb H_0$. The selection of $\mu$ is largely based on trial and error, and still, the convergence may be slow. However, it is possible to precondition such a linear system to accelerate its convergence remarkably as shown in the next section. 

\section{Preconditioned SGD}

A preconditioned SGD is defined by
\begin{equation}\label{p_sgd}
	\pmb \theta^{[\rm new]}  =  \pmb \theta^{[\rm old]} - \mu \pmb P \hat{\pmb g}(\pmb \theta^{[\rm old]}) ,
\end{equation}
where $\pmb P$ is a conformable matrix called preconditioner. The SGD in (\ref{sdg_theta}) is a special case of (\ref{p_sgd}) with $\pmb P =\pmb I$, and we should call it the plain SGD. Noting that our target is to minimize the cost function $f(\pmb \theta)$, $\pmb P$ must be positive definite, and symmetric as well by convention, such that the search direction always points to a descent direction. In convex optimization, inverse of the Hessian is a popular preconditioner. However, in general, the Hessian is not easy to obtain, not always positive definite, and for SGD, such an inverse of Hessian preconditioner may significantly amplify the gradient noise, especially when the Hessian is ill-conditioned. In this section, we detail the behaviors of preconditioned SGD in a  general setting where the problem can be non-convex, and $\pmb P$ is not necessarily related to the Hessian.   

\subsection{Convergence of Preconditioned SGD}

We consider the same second order approximation given in (\ref{soa}). With preconditioning, the learning rule in (\ref{iteration_no_P}) becomes   
\begin{equation}\label{iteration_with_P}
	\pmb \theta^{[\rm new]} = (\pmb I - \mu \pmb P\pmb H_0) \pmb \theta^{[\rm old]} - \mu\pmb  P(\pmb g_0 - \pmb H_0 \pmb \theta_0 + \pmb \epsilon).
\end{equation}      
To proceed, let us first prove the following statement. 

{\it Proposition 1:} All the eigenvalues of $\pmb P \pmb H_0$ are real, and for nonzero eigenvalues, their signs are the same as those of the eigenvalues of $\pmb H_0$.

{\it Proof:} Let $\pmb P^{0.5}$ denote the principal square root of $\pmb P$. Since $\pmb P$ is positive definite, $\pmb P^{0.5}$ is symmetric and positive definite as well. First, we show that $\pmb P\pmb H_0$ and $\pmb P^{0.5}\pmb H_0\pmb P^{0.5}$ have the same eigenvalues. Supposing $\pmb v$ is an eigenvector of $\pmb P^{0.5}\pmb H_0\pmb P^{0.5}$  associated with eigenvalue $\lambda$, i.e., $\pmb P^{0.5}\pmb H_0\pmb P^{0.5} \pmb v = \lambda \pmb v$, then we have $$\pmb P\pmb H_0(\pmb P^{0.5}\pmb v) = \pmb P^{0.5}\pmb P^{0.5}\pmb H_0\pmb P^{0.5} \pmb v = \lambda \pmb P^{0.5}\pmb v.$$ 
As $\pmb P^{0.5}$ is positive definite, $\pmb P^{0.5}\pmb v \ne \pmb 0$. Thus $\pmb P^{0.5}\pmb v$ is an eigenvector of $\pmb P\pmb H_0$ associated with eigenvalue $\lambda$. Similarly, if $\pmb v$ is an eigenvector of $\pmb P\pmb H_0$ associated with eigenvalue $\lambda$, then by rewriting $\pmb P\pmb H_0 \pmb v = \lambda \pmb v$ as
\[ (\pmb P^{0.5}\pmb H_0 \pmb P^{0.5}) \pmb P^{-0.5}\pmb v = \lambda \pmb P^{-0.5}\pmb v, \]
we see that $\lambda$ is an eigenvalue of $\pmb P^{0.5}\pmb H_0 \pmb P^{0.5}$ as well. Hence $\pmb P\pmb H_0$ and $\pmb P^{0.5}\pmb H_0\pmb P^{0.5}$ have identical eigenvalues, and they are real as $\pmb P^{0.5}\pmb H_0\pmb P^{0.5}$ is symmetric. Second, matrices $\pmb P^{0.5}\pmb H_0\pmb P^{0.5}$ and $\pmb H_0$ are congruent, and thus their eigenvalues have the same signs, which implies that the eigenvalues of $\pmb P\pmb H_0$ and $\pmb H_0$ have the same signs as well. \hfill $\square$

Basically, Proposition 1 states that a preconditioner does not change the local geometric structure around $\pmb \theta_0$ in the sense that a local minimum, or maximum, or saddle point before preconditioning keeps to be a local minimum, or maximum, or saddle point after preconditioning, as shown in a numerical example given in Fig.~1.

By introducing eigenvalue decomposition 
$$\pmb P\pmb H_0 = \pmb V\pmb D\pmb V^{-1},$$
we can rewrite (\ref{iteration_with_P})  as
\begin{equation}\label{iteration_decouple}
	\pmb \vartheta^{[\rm new]} = (\pmb I - \mu \pmb D) \pmb \vartheta^{[\rm old]} - \mu \pmb V^{-1}\pmb P(\pmb g_0 - \pmb H_0 \pmb \theta_0 + \pmb \epsilon),
\end{equation}     
where diagonal matrix $\pmb D$ and nonsingular matrix $\pmb V$ contain the eigenvalues and eigenvectors of $\pmb P\pmb H_0$ respectively, and $\pmb \vartheta = \pmb V^{-1}\pmb \theta$ defines a new parameter vector in the transformed coordinates. Since $\pmb I - \mu \pmb D$ is diagonal,  (\ref{iteration_decouple}) suggests that each dimension of $\pmb \vartheta$ evolves independently. This greatly simplifies the study of preconditioned SGD. Without loss of generality, we consider the $i$th dimension of $\pmb \vartheta$,
\begin{equation}\label{iteration_decouple_i}
	\vartheta^{[\rm new]}_i = (1 - \mu d_i) \vartheta^{[\rm old]}_i - \mu h_i - \mu v_i,
\end{equation}
where $\vartheta_i$ and $h_i$ are the $i$th elements of $\pmb \vartheta$ and $\pmb V^{-1}\pmb P(\pmb g_0 - \pmb H_0 \pmb \theta_0)$ respectively, $d_i$ is the $i$th diagonal element of $\pmb D$, and random variable $v_i$ is the $i$th element of random vector $\pmb V^{-1}\pmb P\pmb \epsilon$. We consider the following three cases. 

{\it $d_i>0$)}: By choosing $0<\mu<{1}/{d_i}$, we have $0<1-\mu d_i < 1$. Then repeatedly applying (\ref{iteration_decouple_i}) will let the expectation of $\vartheta_i$ converge to $-{h_i}/{d_i}$. When the variance of $v_i$ is bounded, the variance of $\vartheta_i$ is bounded  as well.  

{\it $d_i<0$)}: For any step size $\mu>0$, we have $1-\mu d_i > 1$. Iteration (\ref{iteration_decouple_i}) always pushes $\vartheta_i$ away from ${h_i}/{d_i}$.

{\it $d_i=0$)}: This should be a transient state since $\vartheta_i$ always drifts away from such a singular point due to gradient noise. 

Similar pictures hold true on the convergence of $\pmb \vartheta$ as well. When $\pmb H_0$ is positive definite, the diagonal elements of $\pmb D$ are positive according to Proposition 1. Then by choosing $0<\mu < {1}/{\max d_i}$, repeated applications of iteration (\ref{iteration_decouple}) let the expectation of $\pmb \theta$ converge to $\pmb \theta_0 - \pmb H_0^{-1} \pmb g_0$, a local minimum of $f(\pmb \theta)$ around $\pmb \theta_0$, where $\max d_i$ denotes the maximum eigenvalue. When $\pmb H_0$ is negative definite, $\pmb \theta_0$ is located around a local maximum of $f(\pmb \theta)$, and iteration (\ref{iteration_decouple}) pushes the expectation of $\pmb \theta$ away from $\pmb \theta_0- \pmb H_0^{-1}\pmb g_0$. When $\pmb H_0$ has both positive and negative eigenvalues, by choosing $0<\mu < {1}/{\max d_i}$, the part of $\pmb \vartheta$ associated with positive eigenvalues is attracted to $\pmb V^{-1}\left(\pmb \theta_0- \pmb H_0^{-1}\pmb g_0\right)$, and the part associated with negative eigenvalues is repelled away from $\pmb V^{-1}\left(\pmb \theta_0- \pmb H_0^{-1}\pmb g_0\right)$.  Saddle point is instable due to gradient noise. 

\begin{figure}[h]
	\centering
	\includegraphics[width=0.9\columnwidth]{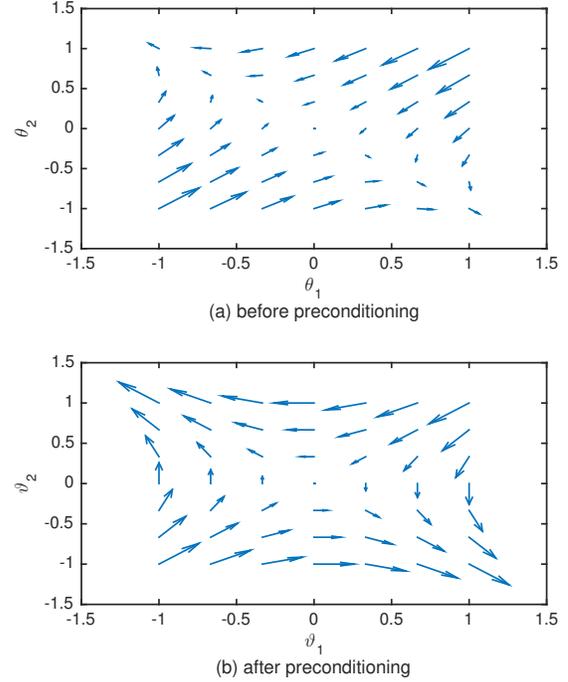}\\
	\caption{ (a) Gradient vectors of quadratic function $-(0.75\theta_1^2+2.5\theta_1\theta_2+0.75\theta_2^2)$. Eigenvalue along the diagonal direction is $0.5$, and $-2$ along the anti-diagonal direction. (b) Preconditioned gradient vectors with preconditioner $[1.25, -0.75; -0.75, 1.25]$. After preconditioning, eigenvalues along the diagonal and anti-diagonal directions do not change their signs, but are scaled to the same amplitude.  }
\end{figure}

\subsection{Three Desirable Properties of a Preconditioner}

We expect a good preconditioner to have the following three desired properties. 

\subsubsection{Small eigenvalue spread} 

In order to achieve approximately uniform convergence rates on all the coordinates of $\pmb \vartheta$, all the eigenvalues of $\pmb P\pmb H_0$ should have similar amplitudes. We use the standard deviation of the logarithms of the absolute eigenvalues of a matrix to measure its eigenvalue spread. The eigenvalue spread gain of a preconditioner $\pmb P$ is defined as the ratio of the eigenvalue spread of $\pmb H_0$ to the eigenvalue spread of $\pmb P\pmb H_0$. Larger eigenvalue spread gain is preferred, and as a base line, a plain SGD has eigenvalue spread gain $1$. 

\subsubsection{Normalized eigenvalue amplitudes}

We hope that all the absolute eigenvalues of $\pmb P\pmb H_0$ are close to $1$ to facilitate the step size selection in preconditioned SGD. Note that in deterministic optimization, the step size can be determined by line search, and thus the scales of eigenvalues are of less interest. However, in stochastic optimization, without the help of line search and knowledge of Hessian, step size selection can be tricky. We use the mean absolute eigenvalues of $\pmb P\pmb H_0$ to measure the scaling effect of $\pmb P$. In a well preconditioned SGD, a normalized step size, i.e., $0<\mu<1$, should work well for the whole learning process, eliminating any manual step size tweaking effort. 

\subsubsection{Large stochastic gradient noise suppression gain} 

Unlike the deterministic optimization, preconditioning for SGD comes at a price, amplification of gradient noise. For the plain SGD in (\ref{sdg_theta}), the gradient noise energy is $ (\pmb \epsilon')^T \pmb \epsilon'$; while for the preconditioned SGD in (\ref{p_sgd}), this noise energy is $ (\pmb \epsilon')^T \pmb P^2 \pmb \epsilon'$. We use the preconditioned gradient noise energy of Newton method, which is $ (\pmb \epsilon')^T \pmb H_0^{-2} \pmb \epsilon'$, as the reference, and define the noise suppression gain of preconditioner $\pmb P$ as
\[
	\frac{E \left[ (\pmb \epsilon')^T \pmb H_0^{-2} \pmb \epsilon' \right] }{E\left[ (\pmb \epsilon')^T \pmb P^2 \pmb \epsilon'\right] } = \frac{ {\rm tr}\left\{ \pmb H_0^{-2} E \left[ \pmb \epsilon' (\pmb \epsilon')^T   \right] \right\} }{ {\rm tr}\left\{ \pmb P^{2} E \left[ \pmb \epsilon' (\pmb \epsilon')^T   \right] \right\} },
\]
where $\rm tr(\cdot)$ takes the trace of a matrix. A good approximation of noise suppression gain is ${\rm tr}(\pmb H_0^{-2})/{\rm tr}(\pmb P^{2})$. 

Unfortunately, due to the existence of gradient noise, generally we cannot find a single preconditioner that simultaneously satisfies all our expectations. When the gradient noise vanishes, for nonsingular $\pmb H_0$, we indeed can find at least one ideal preconditioner such that all the eigenvalues of $\pmb P\pmb H_0$ have unitary amplitude, as stated in the following proposition.  

{\it Proposition 2:} For any nonsingular symmetric $\pmb H_0$, there exists at least one preconditioner $\pmb P$ such that all the absolute eigenvalues of $\pmb P\pmb H_0$ are unitary, and such a $\pmb P$ is unique when $\pmb H_0$ is positive or negative definite. 

{\it Proof: }First, we show that such a preconditioner exists. Assuming the eigenvalue decomposition of $\pmb H_0$ is $\pmb H_0 = \pmb U_{0}\pmb D_0\pmb U_0^T$, we can construct a desired preconditioner as $\pmb P =\pmb  U_{0}|\pmb D_0^{-1}| \pmb U_0^T$, where $\pmb U_0$ is an orthogonal matrix, $\pmb D_0$ is a diagonal matrix, and $|\cdot|$ takes the element-wise absolute value.

Second, we show that such a preconditioner is unique when $\pmb H_0$ is positive or negative definite. When $\pmb H_0$ is positive definite, according to Proposition 1, $\pmb P\pmb H_0$ has positive eigenvalues. If all these eigenvalues are $1$,  then $\pmb P\pmb H_0=\pmb I$, i.e., $\pmb P=\pmb H_0^{-1}$. For negative definite $\pmb H_0$, similarly we can show that $\pmb P=-\pmb H_0^{-1}$. \hfill $\square$

However, such a preconditioner is not necessarily unique for an indefinite Hessian. For example, for Hessian matrix
\[ \pmb H_0 = \left[
\begin{array}{cc}
1& 0   \\
0 &   -1
\end{array}
\right],  \]
any preconditioner having form 
\[ \pmb P = \frac{1}{\alpha^2 - \beta^2} \left[
\begin{array}{cc}
\alpha^2 + \beta^2 & 2\alpha\beta    \\
2\alpha\beta &   \alpha^2 + \beta^2
\end{array}
\right], \quad |\alpha|\ne |\beta| \]
makes $\pmb P\pmb H_0$ have unitary absolute eigenvalues.  

\section{Preconditioner Estimation Criteria}

In practice, the gradient may be easily evaluated, but not for the Hessian matrix. Thus we focus on the preconditioner estimation methods only using the noisy stochastic gradient information. We first discuss two criteria based on secant equation fitting. Although they are not ideal for non-convex optimization, it is still beneficial to study them in detail as they are intimately related to the deterministic and stochastic quasi-Newton methods. We then propose a new preconditioner estimation criterion suitable for both convex and non-convex stochastic optimizations, and show how it overcomes the fundamental limitations of secant equation fitting based solutions. 

\subsection{Criteria Based on Secant Equation Fitting}

Let $\delta \pmb \theta $ be a small perturbation of $\pmb \theta$ around $\pmb \theta_0$. From (\ref{new_hat_g}), we know that 
\begin{equation}\label{secant_eq0}
	\hat{\pmb g}(\pmb \theta + \delta \pmb \theta) - \hat{\pmb g}(\pmb \theta) = \pmb H_0 \delta \pmb \theta + \pmb \varepsilon,
\end{equation}
where $\pmb \varepsilon$ is a random vector accounting for the errors introduced by stochastic approximation of gradients and second order approximation of cost function. It is proposed to use the same randomly sampled training data to calculate $\hat{\pmb g}(\pmb \theta + \delta \pmb \theta)$ and $ \hat{\pmb g}(\pmb \theta)$ for two reasons. First, this practice reduces stochastic noise, and makes sure that $\hat{\pmb g}(\pmb \theta + \delta \pmb \theta)- \hat{\pmb g}(\pmb \theta)\rightarrow\pmb 0$ when $\delta \pmb \theta\rightarrow \pmb 0$. Second, it avoids reloading or regenerating training data. To simplify the notation, we rewrite (\ref{secant_eq0}) as
\begin{equation}\label{secant_eq}
	\delta \hat{\pmb g} = \pmb H_0 \delta \pmb \theta + \pmb \varepsilon,
\end{equation}
where $\delta \hat{\pmb g} = \hat{\pmb g}(\pmb \theta + \delta \pmb \theta) - \hat{\pmb g}(\pmb \theta) $ denotes a random perturbation of stochastic gradient. We call (\ref{secant_eq}) the stochastic secant equation. In quasi-Newton methods, the secant equation is used to derive diverse forms of estimators for the Hessian or its inverse, and Broyden-Fletcher-Goldfarb-Shanno (BFGS) formula is among the widely used ones. In deterministic optimization, BFGS is used along with line search to ensure that the updated Hessian estimate is always positive definite.  However, in stochastic optimization, due to the existence of gradient noise and the infeasibility of line search, attempts in this direction can only achieve limited successes.  One common assumption to justify the use of (\ref{secant_eq}) for preconditioner estimation is that the true Hessians around $\pmb \theta_0$ are positive definite, i.e., the optimization problem is convex or $\pmb \theta$ already is in a basin of attraction. This can be a serious limitation in practice, although it greatly simplifies the design of stochastic quasi-Newton methods. Nevertheless, secant equation based Hessian estimators are still widely adopted in both deterministic and stochastic optimizations. 

\subsubsection{Criterion 1}

With sufficient independent pairs of $(\delta \pmb \theta, \delta \pmb g)$ around $\pmb \theta_0$, we will be able to estimate $\pmb H_0$ by fitting the secant equation (\ref{secant_eq}). It is natural to assume that the error $\pmb \varepsilon$ is Gaussian distributed, and thus a reasonable criterion for preconditioner estimation is  
\begin{equation}
	c_1(\pmb P) = E\left[ \left\| \delta \hat{\pmb g} - \pmb P^{-1} \delta \pmb \theta \right\|^2 \right],
\end{equation}
where $\|\cdot\|$ denotes the Euclidean length of a vector, $\delta \pmb \theta$ and the associated $\delta \hat{\pmb g}$ are regarded as random vectors, and $E$ takes expectation over them. The preconditioner determined by this criterion is called preconditioner 1. Using equation
\begin{equation}\label{dP-1}
	{\rm d}\pmb P^{-1} = -\pmb P^{-1} {\rm d}\pmb P \pmb P^{-1},
\end{equation}
we can show that the derivative of $c_1(\pmb P)$ with respect to $\pmb P$ is
\begin{equation}
	\frac{\partial c_1(\pmb P)}{\partial \pmb P} = \pmb P^{-1} \left( \pmb e_1 \delta \pmb \theta^T  + \delta \pmb \theta \pmb e_1^T \right) \pmb P^{-1},
\end{equation}
where ${\rm d}$ denotes differentiation, and $\pmb e_1 = \delta \hat{\pmb g} - \pmb P^{-1} \delta \pmb \theta$. 
Noting that $\pmb P$ is symmetric, the gradient of $c_1$ with respect to $\pmb P$ is symmetric as well, or equivalently, the symmetric part of the gradient of $c_1$ with respect to $\pmb P$ without considering symmetry constraint.
By letting the gradient be zero, we can find the optimal $\pmb P$ by solving equation
\begin{equation}\label{lyap_H1}
	\pmb R_{\theta} \pmb P^{-1} + \pmb P^{-1} \pmb R_{\theta} - (\pmb R_{\theta g} + \pmb R_{g\theta }) = \pmb 0,
\end{equation}
where $\pmb R_{\theta} = E[\delta \pmb \theta \delta \pmb \theta^T]$, $\pmb R_{\theta g} = E[\delta \pmb \theta \delta \hat{\pmb g}^T]$, and $\pmb R_{g\theta} = E[\delta \hat{\pmb g} \delta \pmb \theta^T]$.

Equation (\ref{lyap_H1}) is a continuous Lyapunov equation well known in control theory. Using result
\begin{equation}\label{vec_ABC}
	{\rm vec}(\pmb A\pmb B\pmb C) = (\pmb C^T\otimes \pmb A){\rm vec}(\pmb B),
\end{equation}
we can rewrite (\ref{lyap_H1}) as
\begin{equation}\label{solution_lyap_H1}
	( \pmb I \otimes \pmb R_{\theta} + \pmb R_{\theta} \otimes \pmb I ) {\rm vec}(\pmb P^{-1}) = {\rm vec}(\pmb R_{\theta g} + \pmb R_{g\theta})
\end{equation}
to solve for $\pmb P$, where ${\rm vec}(\pmb A)$ is the vector formed by stacking the columns of $\pmb A$, $\otimes$ denotes matrix Kronecker product, and $\pmb I$ is a conformable identity matrix.

However, the solution given by (\ref{solution_lyap_H1}) is not intuitive. We can solve for $\pmb P$ directly with the following mild assumptions:
\begin{itemize}
	\item[A1)] $\delta \pmb \theta$ has zero mean, i.e., $E[\delta \pmb \theta] = \pmb 0$.
	\item[A2)] $\delta \pmb \theta$ and $\pmb \varepsilon$ are uncorrelated, i.e., $E[\delta \pmb \theta \pmb \varepsilon^T] = \pmb 0$.
	\item[A3)] Covariance matrix $\pmb R_{\theta} = E[\delta \pmb \theta \delta \pmb \theta^T]$ is positive definite. 
\end{itemize}
With the above assumptions, we have 
\begin{equation}\label{R_gx}
	\pmb R_{g\theta } = \pmb H_0 \pmb R_{\theta}, \quad \pmb R_{\theta g} = \pmb R_{\theta} \pmb H_0.
\end{equation}
Using (\ref{R_gx}), (\ref{lyap_H1}) becomes
\begin{equation}
	\pmb R_{\theta} (\pmb P^{-1} - \pmb H_0) + (\pmb P^{-1}-\pmb H_0) \pmb R_{\theta} = \pmb 0,
\end{equation}
or equivalently 
\begin{equation}\label{P1=H0}
	( \pmb I \otimes \pmb R_{\theta} + \pmb R_{\theta} \otimes \pmb I ) {\rm vec}(\pmb P^{-1} - \pmb H_0) = \pmb 0
\end{equation}
by using (\ref{vec_ABC}).
Since $\pmb R_{\theta} $ is positive definite, $\pmb I \otimes \pmb R_{\theta} + \pmb R_{\theta} \otimes \pmb I $ is positive definite as well. Hence ${\rm vec}(\pmb P^{-1} - \pmb H_0)=\pmb 0$ by (\ref{P1=H0}), i.e., $\pmb P=\pmb H_0^{-1}$. Thus as in the deterministic optimization, with enough independent pairs of $(\delta \pmb \theta, \delta \pmb g)$, criterion 1 leads to an asymptotically unbiased estimation of $\pmb H_0^{-1}$. Such an unbiasedness property is preferred in deterministic optimization, however, it might be undesirable in stochastic optimization as such a preconditioner may significantly amplify the gradient noise. Furthermore, when preconditioner 1 is estimated using finite pairs of $(\delta \pmb \theta, \delta \pmb g)$, $\pmb P$ cannot be guaranteed to be positive definite even if $\pmb H_0$ is positive definite.

\subsubsection{Criterion 2}

By rewriting (\ref{secant_eq}) as $\pmb H_0^{-1}\delta \hat{\pmb g} =  \delta \pmb \theta + \pmb H_0^{-1} \pmb \varepsilon$, we may introduce another criterion for secant equation fitting,
\begin{equation}
	c_2(\pmb P) = E [ \| \pmb P \delta \hat{\pmb g} - \delta \pmb \theta \|^2 ].
\end{equation}
Naturally, the preconditioner determined by this criterion is called preconditioner 2.
The derivative of $c_2$ with respect to $\pmb P$ is
\begin{equation}
	\frac{\partial c_2(\pmb P)}{\partial \pmb P} = \delta \hat{\pmb g} \pmb e_2^T + \pmb e_2  \delta \hat{\pmb g}^T,
\end{equation}
where $\pmb e_2 = \pmb P \delta \hat{\pmb g} - \delta \pmb \theta$. By letting the derivative of $c_2$ with respect to $\pmb P$ be zero, we find that the optimal $\pmb P$ satisfies equation
\begin{equation}\label{another_lyap}
	\pmb P \pmb R_g + \pmb R_g \pmb P - \pmb R_{g\theta} - \pmb R_{\theta g} = \pmb 0,
\end{equation}
where $\pmb R_g = E[\delta \hat{\pmb g} \delta \hat{\pmb g}^T]$. 

Again, (\ref{another_lyap}) is a continuous Lyapunov equation, and can be numerically solved.  Unfortunately, there does not exist a simple closed-form relationship between the optimal $\pmb P$ and $\pmb H_0$. Still, analytical solutions in simplified scenarios can cast crucial insight into the properties of this criterion. By assuming assumption A4,
\begin{equation}\label{white_R}
	\pmb R_{\theta} = \sigma^2_{\theta} \pmb I, \quad \pmb R_{\varepsilon} = E[\pmb \varepsilon\pmb \varepsilon^T]=\sigma^2_{\varepsilon} \pmb I,
\end{equation}
the optimal $\pmb P$ can be shown to be
\begin{equation}\label{P2_close_form}
	\pmb P = \sum_{i=1}^m \frac{\lambda_i}{\lambda_i^2 + {\sigma^2_{\varepsilon}}/{\sigma^2_{\theta}}} \pmb u_i \pmb u_i^T,
\end{equation}
where $\pmb H_0 = \sum_{i=1}^m \lambda_i \pmb u_i \pmb u_i^T$ is the eigenvalue decomposition of $\pmb H_0$ with $\pmb u_i$ being its $i$th eigenvector associated with eigenvalue $\lambda_i$. As criterion 1, criterion 2 leads to an unbiased estimation of $\pmb H_0^{-1}$ when $\sigma^2_{\varepsilon}=0$. But unlike criterion 1, the optimal $\pmb P$ here underestimates the inverse of Hessian when $\sigma^2_{\varepsilon}>0$. Such a built-in annealing mechanism is actually desired in stochastic optimization. Again, when preconditioner 2 is estimated using finite pairs of $(\delta \pmb \theta, \delta \pmb g)$, it can be indefinite even if $\pmb H_0$ is positive definite. 

\subsection{A New Criterion for Preconditioner Estimation}   

As examined in Section III, the essential utility of a preconditioner is to ensure that $\pmb \vartheta$ enjoys approximately uniform convergence rates across all directions. Amplitudes, but not the signs, of the eigenvalues of Hessian are to be normalized. There is no need to explicitly estimate the Hessian. We propose the following new criterion for preconditioner estimation,
\begin{equation}\label{c3}
	c_3(\pmb P) = E[ \delta \hat{\pmb g}^T \pmb P \delta \hat{\pmb g} + \delta \pmb \theta^T \pmb P^{-1} \delta \pmb \theta],
\end{equation}
and call it and resultant preconditioner criterion 3 and preconditioner 3 respectively. The rationale behind criterion 3 is that when $c_3(\pmb P)$ is minimized, we should have 
\[ \left. \frac{\partial c_3(\kappa \pmb P)}{\partial \kappa}\right|_{\kappa=1} = 0, \]
which suggests its two terms, $E[ \delta \hat{\pmb g}^T \pmb P \delta \hat{\pmb g} ]$ and $E[ \delta \pmb \theta^T \pmb P^{-1} \delta \pmb \theta]$, are equal, and thus the amplitudes of perturbations of preconditioned stochastic gradients match that of parameter perturbations as in the Newton method where the inverse of Hessian is the preconditioner. Furthermore, $c_3(\pmb P)$ is invariant to the changes of the signs of $\delta \hat{\pmb g}$ and $\delta \pmb \theta$, i.e., pairs $(\pm \delta \pmb \theta, \pm \delta \pmb g)$ resulting in the same cost. We present the following proposition to rigorously justify the use of criterion 3. To begin with, we first prove a lemma. 

{\it Lemma 1:} If $\pmb A$ is symmetric, $\pmb A^2 = \pmb D$, and $\pmb D$ is a diagonal matrix with distinct nonnegative diagonal elements, then we have $\pmb A = \pmb D_{\rm sign}{\pmb D}^{0.5}$, where $\pmb D_{\rm sign} $ is an arbitrary diagonal matrix with diagonal elements being either $1$ or $-1$. 

{\it Proof:} Let the eigenvalue decomposition of $\pmb A$ be $\pmb A=\pmb U \pmb D_A \pmb U^T$. Then $\pmb A^2 = \pmb D$ suggests $\pmb U \pmb D_A^2 \pmb U^T = \pmb D$. Noting that the eigenvalue decomposition of $\pmb D$ is unique when it has distinct diagonal elements, we must have $\pmb U=\pmb I$ and $\pmb D_A^2 = \pmb D$, i.e., $\pmb A = \pmb D_{\rm sign}{\pmb D}^{0.5}$.  \hfill $\square$

{\it Proposition 3:} For positive definite covariance matrices $ \pmb R_{\theta}$ and $\pmb R_g$, criterion $c_3( \pmb P)$ determines an optimal positive definite preconditioner $ \pmb P$ scaling $\delta \hat{ \pmb g}$ as $ \pmb PE[\delta \hat{ \pmb g}\delta \hat{ \pmb g}^T] \pmb P=E[\delta  \pmb \theta\delta  \pmb \theta^T]$. The optimal $\pmb P$ is unique when $\pmb R_{\theta}^{0.5} \pmb R_g \pmb R_{\theta}^{0.5}$ has distinct eigenvalues. 

{\it Proof:} The derivative of $c_3( \pmb P)$ with respect to $ \pmb P$ is 
\begin{equation}
	\frac{\partial c_3( \pmb P)}{\partial  \pmb P} = E[\delta \hat{ \pmb g}\delta \hat{ \pmb g}^T] -  \pmb P^{-1}E[\delta  \pmb \theta\delta  \pmb \theta^T] \pmb P^{-1}.
\end{equation}
By letting the gradient be zero, we obtain the following equation for optimal $ \pmb P$,
\begin{equation}\label{Riccati}
	 \pmb P \pmb R_g \pmb P -   \pmb R_{\theta} = \pmb 0,
\end{equation}
which has the form of a continuous time algebraic Riccati equation known in control theory, but lacking the linear term of $ \pmb P$. To solve for $ \pmb P$, we rewrite (\ref{Riccati}) as
\begin{equation}\label{care1}
	( \pmb R_{\theta}^{-0.5} \pmb P \pmb R_{\theta}^{-0.5} ) \pmb R_{\theta}^{0.5} \pmb R_g \pmb R_{\theta}^{0.5} (\pmb R_{\theta}^{-0.5} \pmb P \pmb R_{\theta}^{-0.5} ) = \pmb I,
\end{equation}
where $\pmb A^{0.5}$ denotes the principal square root of a positive definite matrix $\pmb A$. By introducing eigenvalue decomposition
\begin{equation}
	\pmb R_{\theta}^{0.5} \pmb R_g \pmb R_{\theta}^{0.5} = \pmb U \pmb D \pmb U^T,
\end{equation}
we can rewrite (\ref{care1}) as
\begin{equation}
	( \pmb U^T \pmb R_{\theta}^{-0.5} \pmb P \pmb R_{\theta}^{-0.5} \pmb U ) \pmb D ( \pmb U^T \pmb R_{\theta}^{-0.5} \pmb P \pmb R_{\theta}^{-0.5} \pmb U )  = \pmb I,
\end{equation}
or equivalently,
\begin{equation}
 \pmb D   = ( \pmb U^T \pmb R_{\theta}^{-0.5} \pmb P \pmb R_{\theta}^{-0.5} \pmb U )^{-2}.
\end{equation}
When $\pmb R_{\theta}^{0.5} \pmb R_g \pmb R_{\theta}^{0.5}$ does not have repeated eigenvalues, the diagonal elements of $\pmb D$ are distinct, and the solution $\pmb P$ must have form
\begin{equation}\label{P_solution}
	\pmb P = \pmb R_{\theta}^{0.5}  \pmb U { ( \pmb D_{\rm sign} \pmb D^{-0.5}  )} \pmb U^T \pmb R_{\theta}^{0.5}
\end{equation}
by Lemma 1. For positive definite $\pmb P$, we can only choose $\pmb D_{\rm sign} = \pmb I$, and thus the optimal $\pmb P$ is unique. When $\pmb R_{\theta}^{0.5} \pmb R_g \pmb R_{\theta}^{0.5}$ has repeated eigenvalues, (\ref{P_solution}) still gives a valid solution, but not necessarily the only one. The assumption of non-singularity of  $\pmb R_{\theta}$ and $\pmb R_g$ is used to make the involved matrix inversion feasible. \hfill $\square$

Unlike the two secant equation fitting based preconditioners, preconditioner 3 is guaranteed to be positive definite as long as the estimated covariance matrices $\hat{\pmb R}_\theta$ and $\hat{\pmb R}_g$ are positive definite. To gain more insight into the properties of preconditioner 3, we consider closed-form solutions in simplified scenarios that can explicitly link the optimal $\pmb P$ to $\pmb H_0$.  When $\pmb R_{\theta}$ and $\pmb R_{\varepsilon}$ have the simple forms shown in (\ref{white_R}), the closed-form solution for $\pmb P$ is 
\begin{equation}
	\pmb P = \sum_{i=1}^m \frac{1}{ \sqrt{ \lambda_i^2 + {\sigma^2_{\varepsilon}}/{\sigma^2_{\theta}} }}  \pmb u_i \pmb u_i^T,
\end{equation}
where $\pmb H_0 = \sum_{i=1}^m \lambda_i \pmb u_i \pmb u_i^T$ is the eigenvalue decomposition of $\pmb H_0$. The eigenvalues of $\pmb P\pmb H_0$ are $ {\lambda_i}/{ \sqrt{ \lambda_i^2 + {\sigma^2_{\varepsilon}}/{\sigma^2_{\theta}} }}, \; 1\le i\le m$. When $\sigma^2_{\varepsilon}\rightarrow \infty$, we have
\[  \frac{\lambda_i}{ \sqrt{ \lambda_i^2 + {\sigma^2_{\varepsilon}}/{\sigma^2_{\theta}} }} \rightarrow \frac{\sigma_{\theta}\lambda_i}{\sigma_{\varepsilon} }.\]
Thus for heavily noisy gradient, the optimal preconditioner cannot improve the eigenvalue spread, but are damping the gradient noise. Such a built-in step size adjusting mechanism is highly desired in stochastic optimization.  When $\sigma^2_{\varepsilon}=0$, $\pmb P$ reduces to the ideal preconditioner constructed in the proof of Proposition 2, and all the eigenvalues of $\pmb P\pmb H_0$ have unitary amplitude. These properties make preconditioner 3 an ideal choice for preconditioned SGD. 

\subsection{Relationship to Newton Method}

In a Newton method for deterministic optimization, we have $\delta \pmb g = \pmb H_0 \delta \pmb \theta$. Here $\delta \pmb g$ is an exact gradient perturbation. Let us rewrite this secant equation in matrix form as $\pmb H_0^{-1} \delta \pmb g\delta \pmb g^T \pmb H_0^{-1} = \delta \pmb \theta \delta \pmb \theta^T $, and compare it with relation $ \pmb PE[\delta \hat{ \pmb g}\delta \hat{ \pmb g}^T] \pmb P=E[\delta  \pmb \theta\delta  \pmb \theta^T]$ from Proposition 3. Now it is clear that preconditioner 3 scales the stochastic gradient in a way comparable to Newton method in deterministic optimization.    

The other two preconditioners either over or under compensate the stochastic gradients, inclining to cause divergence or slowdown convergence as shown in our experimental results. To simplify our analysis work, we consider the closed-form solutions of the first two preconditioners. For preconditioner 1, we have $\pmb P = \pmb H_0^{-1}$ under assumptions A1, A2 and A3. Then with (\ref{secant_eq}), we have
\[
	\pmb PE[\delta \hat{ \pmb g}\delta \hat{ \pmb g}^T] \pmb P - E[\delta  \pmb \theta\delta  \pmb \theta^T] = \pmb H_0^{-1} E[\pmb \varepsilon\pmb \varepsilon^T] \pmb H_0^{-1} \succeq \pmb 0,
\]
which suggests that preconditioner 1 over compensates the stochastic gradient, where $\succeq \pmb 0$ means that the matrix on the left side of $\succeq$ is positive semidefinite definite. For preconditioner 2, using the closed-form solution in (\ref{P2_close_form}), we have
\[
\pmb PE[\delta \hat{ \pmb g}\delta \hat{ \pmb g}^T] \pmb P - E[\delta  \pmb \theta\delta  \pmb \theta^T] = -\sum_{i=1}^m \frac{\sigma^2_{\varepsilon} \pmb u_i \pmb u_i^T}{\lambda_i^2 + {\sigma^2_{\varepsilon}}/{\sigma^2_{\theta}}} \preceq\pmb 0,
\]
which suggests that preconditioner 2 under compensates the stochastic gradient, where $\preceq \pmb 0$ means that the matrix on the left side of $\preceq$ is negative semidefinite. 

\section{Preconditioner Estimation Methods}

It is possible to design different preconditioner estimation methods based on the criteria proposed in Section IV. To minimize the overhead of preconditioned SGD, we only consider the simplest preconditioner estimation methods: SGD algorithms for learning $\pmb P$ minimizing the criteria in Section IV with mini-batch size $1$. We call the SGD for $\pmb \theta$ the primary SGD to distinguish it from the SGD for learning $\pmb P$. In each iteration of preconditioned SGD, we first evaluate the gradient twice to obtain $\hat{\pmb g}(\pmb \theta)$ and $\hat{\pmb g}(\pmb \theta + \delta \pmb \theta)$. Then the pair, $(\delta\pmb  \theta, \delta \hat{\pmb g})$, is used to update the preconditioner estimation once. Lastly, (\ref{p_sgd}) is used to update $\pmb \theta$ to complete one iteration of preconditioned SGD.   

\subsection{Dense Preconditioner}

We focus on the algorithm design for criterion 3. Algorithms for the other two criteria are similar, and will be given without detailed derivation. In this subsection, we do not assume that the preconditioner has any sparse structure except for being symmetric, i.e. it is a dense matrix. Elements in the Hessian, and thus the preconditioner, can have a large dynamic range. Additive learning rules like the regular gradient descent may converge slow when the preconditioner is poorly initialized. We find that multiplicative updates, e.g., the relative (natural) gradient descent, could perform better due to its equivariant property \cite{Amari96,Cardoso96}. However, to use the relative gradient descent, we need to find a Lie group representation for $\pmb P$. It is clear that positive definite matrices do not form a Lie group under matrix multiplication operation. Let us consider the Cholesky factorization
\begin{equation}
	\pmb P = \pmb Q^T \pmb Q,
\end{equation}
where $\pmb Q$ is an upper triangular matrix with positive diagonal elements. It is straightforward to show that all upper triangular matrices with positive diagonal elements form a Lie group under matrix multiplication operation. Thus in order to use the relative gradient descent, we shall learn the matrix $ \pmb Q$.  One desired by-product of Cholesky factorization is that the resultant triangular system can be efficiently solved by forward or backward substitution. 

Following the third criterion, the instantaneous cost to be minimized is
\begin{equation}
	\hat{c}_3(\pmb P) = \delta \hat{\pmb g}^T \pmb P \delta \hat{\pmb g} + \delta \pmb \theta^T \pmb P^{-1} \delta \pmb \theta,
\end{equation}
which is an approximation of (\ref{c3}) with mini-batch size $1$. We consider a small perturbation of  $\pmb Q$ given by $\delta \pmb Q = \mathcal{E} \pmb Q$, where $ \mathcal{E}$ is an infinitely small upper triangle matrix such that $\pmb Q + \delta \pmb Q$ still belongs to the same Lie group. The relative gradient is defined by
\[ \pmb \nabla \mathcal{E} =  \left. \frac{\partial \hat{c}_3(\pmb Q + \mathcal{E} \pmb Q)}{\partial \mathcal{E}} \right|_{\mathcal{E} = \pmb 0}, \]
where with slight abuse of notation, $c_3$ is rewritten as a function of $\pmb Q$.
Using (\ref{dP-1}), we can show that 
\begin{equation}
	  \pmb \nabla \mathcal{E} = 2 {\rm triu}  \left( \pmb Q\delta \hat{\pmb g} \delta\hat{\pmb g}^T \pmb Q^T - \pmb  Q^{-T} \delta \pmb \theta \delta \pmb \theta^T \pmb Q^{-1}\right),
\end{equation}
where operator ${\rm triu}(\cdot)$ takes the upper triangular part of a matrix. Then $\pmb Q$ can be updated as
\begin{equation}
	\pmb Q^{[\rm new]} = \pmb Q^{[\rm old]} - \mu_Q \pmb \nabla \mathcal{E} \pmb Q^{[\rm old]},
\end{equation}
where $\mu_Q >0$ is a small enough step size such that the diagonal elements of $\pmb Q^{[\rm new]}$ keep to be positive. To simplify the step size selection, normalized step size
\begin{equation}\label{Q_mu0}
	\mu_Q = \frac{\mu_{Q,0}}{ \max | \pmb  \nabla \mathcal{E}  |}
\end{equation}
can be used, where $0< \mu_{Q,0} < 1$, and $ \max |  \pmb \nabla \mathcal{E}  |$ denotes the maximum element-wise absolute value of $\pmb \nabla \mathcal{E}$. Another normalized step size 
\[
	\mu_Q = \frac{\mu_{Q,0}}{ \max | {\rm diag} (\pmb  \nabla \mathcal{E} ) |}
\]
can be useful, and also ensures that $\pmb Q^{[\rm new]}$ belongs to the same Lie group when $0< \mu_{Q,0} < 1$, where $ \max | {\rm diag} ( \pmb \nabla \mathcal{E} ) |$ denotes the maximum absolute value of the diagonal elements of $\pmb \nabla \mathcal{E}$. Our experiences suggest that the two step size normalization strategies are comparable in stochastic optimization. However, the one given in (\ref{Q_mu0}) seems to be preferred in deterministic optimization as it leads to more stable convergence.  

We summarize the complete preconditioned SGD as below. 

\hrulefill

{\it One complete iteration of preconditioned SGD with preconditioner 3}

Inputs are $ \pmb \theta^{[\rm old]}$ and $\pmb Q^{[\rm old]}$; outputs are $ \pmb \theta^{[\rm new]}$ and $\pmb Q^{[\rm new]}$.

\hrulefill

\begin{itemize}
	\item [1) ] Sample $\delta \pmb \theta$, and calculate $\hat{\pmb g}(\pmb \theta^{[\rm old]})$ and $\delta \hat{\pmb g} = \hat{\pmb g}(\pmb \theta^{[\rm old]} + \delta \pmb \theta) - \hat{\pmb g}(\pmb \theta^{[\rm old]}) $.
	\item [2) ] Calculate $\pmb a = \pmb Q^{[\rm old]}\delta \hat{\pmb g}$, and $\pmb b=\left(\pmb Q^{[\rm old]} \right)^{-T} \delta \pmb \theta$ via solving triangular system $\left(\pmb Q^{[\rm old]} \right)^{T} \pmb b= \delta \pmb \theta$. 
	\item [3) ] Update the preconditioner by 
	$$\pmb Q^{[\rm new]} = \pmb Q^{[\rm old]} - \frac{\mu_{Q, 0}}{ \max | \pmb \nabla \mathcal{E}  |} \pmb \nabla \mathcal{E} \pmb Q^{[\rm old]},$$
	where $\pmb \nabla \mathcal{E} =  2{\rm triu} (\pmb a \pmb a^T-\pmb b\pmb b^T)$, and $0<\mu_{Q, 0}<1$.
		\item [4) ] Update $\pmb \theta$ by
		\[  \pmb \theta^{[\rm new]} = \pmb \theta^{[\rm old]} - \mu_{\theta, 0} \left(\pmb Q^{[\rm new]}\right)^T \pmb Q^{[\rm new]} \hat{\pmb g}(\pmb \theta^{[\rm old]}), \] 
		where $0<\mu_{\theta, 0}<1$. 
\end{itemize}

\hrulefill

Here, elements of $\delta \pmb \theta$ can be sampled from the Gaussian distribution with a small variance. Then the only tunable parameters are the two normalized step sizes, $\mu_{Q, 0}$ and $\mu_{\theta, 0}$. 

Algorithm design for the other two preconditioners is similar, and we give their relative gradients for updating $\pmb Q$ without derivation. For criterion 1, the relative gradient is 
\[ \pmb \nabla \mathcal{E} =  2 {\rm triu} \left( \pmb Q^{-T} \delta \pmb \theta  \pmb e_1^T \pmb Q^{-1} + \pmb Q^{-T} \pmb e_1 \delta \pmb \theta^T \pmb Q^{-1} \right), \]
where $\pmb e_1 =\delta\hat{\pmb g} -\pmb  P^{-1}\delta\pmb \theta $. For criterion 2, the relative gradient is 
\[  \pmb \nabla \mathcal{E} = 2 {\rm triu}\left( \pmb Q \delta\hat{\pmb g} \pmb e_2^T \pmb Q^T +  \pmb Q \pmb e_2  \delta\hat{\pmb g}^T \pmb Q^T  \right), \]
where $\pmb e_2 = \pmb P\delta\hat{\pmb g} - \delta\pmb \theta$. It is worthy to mention that by writing $\pmb P=\pmb Q^T\pmb Q$ when using criteria 1 and 2, we are forcing the preconditioners to be positive definite, but the optimal solution minimizing criterion 1 or 2 is not necessarily positive definite. For criterion 3, by writing $\pmb P=\pmb Q^T\pmb Q$, we are selecting the positive definite solution among many possible ones. 

\subsection{Preconditioner with Sparse Structures}

In practice, $\pmb \theta$ can have millions of free parameters to learn, and thus a dense $\pmb P$ might have trillions of elements to estimate and store. Clearly, a dense representation for $\pmb P$ is no longer feasible. For large-scale problems, we need to assume that $\pmb P$ has certain sparse structures so that it can be manipulated. One extreme example is to treat $\pmb P$ as a diagonal matrix. However, such a simplification is too coarse, and generally does not lead to significant performance gain over the plain SGD. Application specific knowledge may play the key role in determining a proper form for $\pmb P$. 

One example is to assume $\pmb Q$ has structure
\[ \pmb Q = \left[
\begin{array}{cc}
\pmb Q_{11} & \pmb Q_{12}    \\
\pmb 0 &   \pmb Q_{22}
\end{array}
\right],   \]
where $\pmb Q_{11}$ is an upper triangular matrix, $\pmb Q_{12}$ is a dense matrix, $\pmb Q_{22}$ is a diagonal matrix, and all the diagonal elements of $\pmb Q$ are positive. It is straightforward to show that such limited-memory triangular matrices form a Lie group, and thus relative gradient descent applies here. Cholesky factor of ill-conditioned matrix can be well approximated with this form. However, the dimension of $\pmb Q_{11}$ should be properly selected to achieve a good trade off between representation complexity and accuracy.     

In many situations, the parameters to be optimized naturally form a multi-dimensional array, reflecting certain built-in structures of the training data and the parameter space. Hence we may approximate the preconditioner as Kronecker products of a series of small matrices. For example, preconditioner for a $3\times 4\times 5$ parameter array may be approximated as Kronecker product of three small matrices with sizes $5\times 5$, $4\times 4$ and $3\times 3$. Such a bold simplification often works well in practice. Interestingly, similar ideas have been exploited in \cite{Martens2015, hess_kron}, and achieved certain successes.

To explain the above idea clearly, let us consider a concrete example where the parameters to be optimized naturally form a two dimensional array, i.e., a matrix, denoted by $\pmb \Theta$. Its stochastic gradient, $\hat{\pmb  G}(\pmb \Theta)$, is a matrix with the same dimensions. The preconditioned SGD for updating $\pmb \Theta$ is
\begin{equation}\label{sgd_2d}
	{\rm vec}\left( \pmb \Theta^{[\rm new]} \right) = {\rm vec}\left( \pmb \Theta^{[\rm old]} \right) - \mu \pmb P {\rm vec}\left[ \hat{\pmb G}(\pmb \Theta^{[\rm old]}) \right].
\end{equation}
To simplify the preconditioner estimation, we may assume that $\pmb P =\pmb  P_2\otimes \pmb P_1$, and thus can rewrite (\ref{sgd_2d}) as
\begin{equation}
	\pmb \Theta^{[\rm new]} =  \pmb \Theta^{[\rm old]} - \mu \pmb P_1 \hat{\pmb G}(\pmb \Theta^{[\rm old]}) \pmb P_2
\end{equation}
using (\ref{vec_ABC}), where $\pmb P_1$ and $\pmb P_2$ are two positive definite matrices with conformable dimensions. 
Similarly, by adopting Cholesky factorizations 
\[ \pmb P_1 = \pmb Q_1^T \pmb Q_1, \quad \pmb P_2 = \pmb Q_2^T \pmb Q_2, \]
we can use relative gradients to update $\pmb Q = \pmb Q_2\otimes \pmb Q_1$ as $\delta \pmb Q_1=\mathcal{E}_1 \pmb Q_1$ and $\delta \pmb Q_2 = \mathcal{E}_2 \pmb Q_2$. For criterion 3, the two relative gradients are given by
\begin{eqnarray*}
	\pmb A & =& \pmb Q_1 \delta \hat{\pmb G} \pmb Q_2^T, \\
	\pmb B &=& \pmb Q_2^{-T} \delta\pmb \Theta^T \pmb Q_1^{-1}, \\
	 \pmb \nabla \mathcal{E}_1  & = & 2 {\rm triu} \left(  \pmb A\pmb A^T - \pmb B^T\pmb B \right), \\
 \pmb \nabla \mathcal{E}_2   & = & 2 {\rm triu} \left( \pmb A^T\pmb A - \pmb B\pmb B^T \right) ,
\end{eqnarray*}
where $\pmb B$ can be calculated by solving linear system $\pmb Q_2^T \pmb B \pmb Q_1 = \delta \pmb \Theta^T$ to avoid explicit matrix inversion. The relative gradients for preconditioner updating with criteria 1 and 2 have similar forms, and are not listed here. 

\section{Experimental Results}

Five sets of experimental results are reported in this section. For the preconditioner estimation, we always choose mini-batch size $1$, initialize $\pmb P$ to an identity matrix, set $\mu_{Q,0}=0.01$ in (\ref{Q_mu0}), and sample $\delta \pmb \theta$ from Gaussian distribution $\mathcal{N}(\pmb 0,\; {\rm eps}\times \pmb I)$, where ${\rm eps}=2^{-52}$ is the accuracy in double precision. For the primary SGD, except for the blind equalization example where the mini-batch size is $10$, we always use mini-batch size $100$. Step size for the preconditioned SGD is selected in $[0, 1]$, and in many cases, this range is good for the plain SGD as well. Supplementary materials, including a Matlab code package reproducing all the experimental results here, is available on \url{https://sites.google.com/site/lixilinx/home/psgd}.

\subsection{Criteria and Preconditioners Comparisons}

This experiment compares the three preconditioner estimators developed in Section V. The true Hessian $\pmb H_0$ is a $10\times 10$ symmetric random constant matrix, and its elements are drawn from normal distribution $\mathcal{N}(0, \sigma_h^2)$. We vary three factors for performance study: positive definite Hessian and indefinite Hessian; noise free gradient and heavily noisy gradient with signal-to-noise ratio $-20$ dB; large scale Hessian ($\sigma_h^2 = 10^{12}$) and tiny scale Hessian ($\sigma_h^2 = 10^{-12}$). Totally we have $2^3$ different testing scenarios. Samples of  $\delta\hat{\pmb g}$ and $\delta\pmb\theta$ are generated by model (\ref{secant_eq}), and the task is to estimate a preconditioner with the three desirable properties listed in Section III.B using SGD. 

Fig.~2 shows a set of typical results. We can make the following observations from Fig.~2. Criterion 3 shows robust performance in all the test cases. When the gradient is noise free, it has large eigenvalue spread gains, well normalized eigenvalues, and no amplification of gradient noise. When the gradient is heavily noisy, it cannot improve, but neither worsen, the eigenvalue spread. Instead, it damps the gradient noise, shown by average absolute eigenvalues smaller than $1$ and gradient noise suppression gains larger than $1$. Also, it is worthy to point out that the preconditioner estimation algorithm for criterion 3 shows similar convergence rates when the Hessians have an extremely large numerical dynamic range, a desirable behavior expected due to the equivariant property of relative gradient descent \cite{Cardoso96}. Criteria 1 and 2 fail completely in the cases of indefinite Hessians, and also show limited performance when the Hessians are positive definite. In the third row of Fig.~2, preconditioner 1 does show a larger eigenvalue spread gain than preconditioner 3, but the price is the amplification of gradient noise, as revealed by its smaller gradient noise suppression gain.

\begin{figure*}[!htbp]
	\centering
	\includegraphics[width=\textwidth]{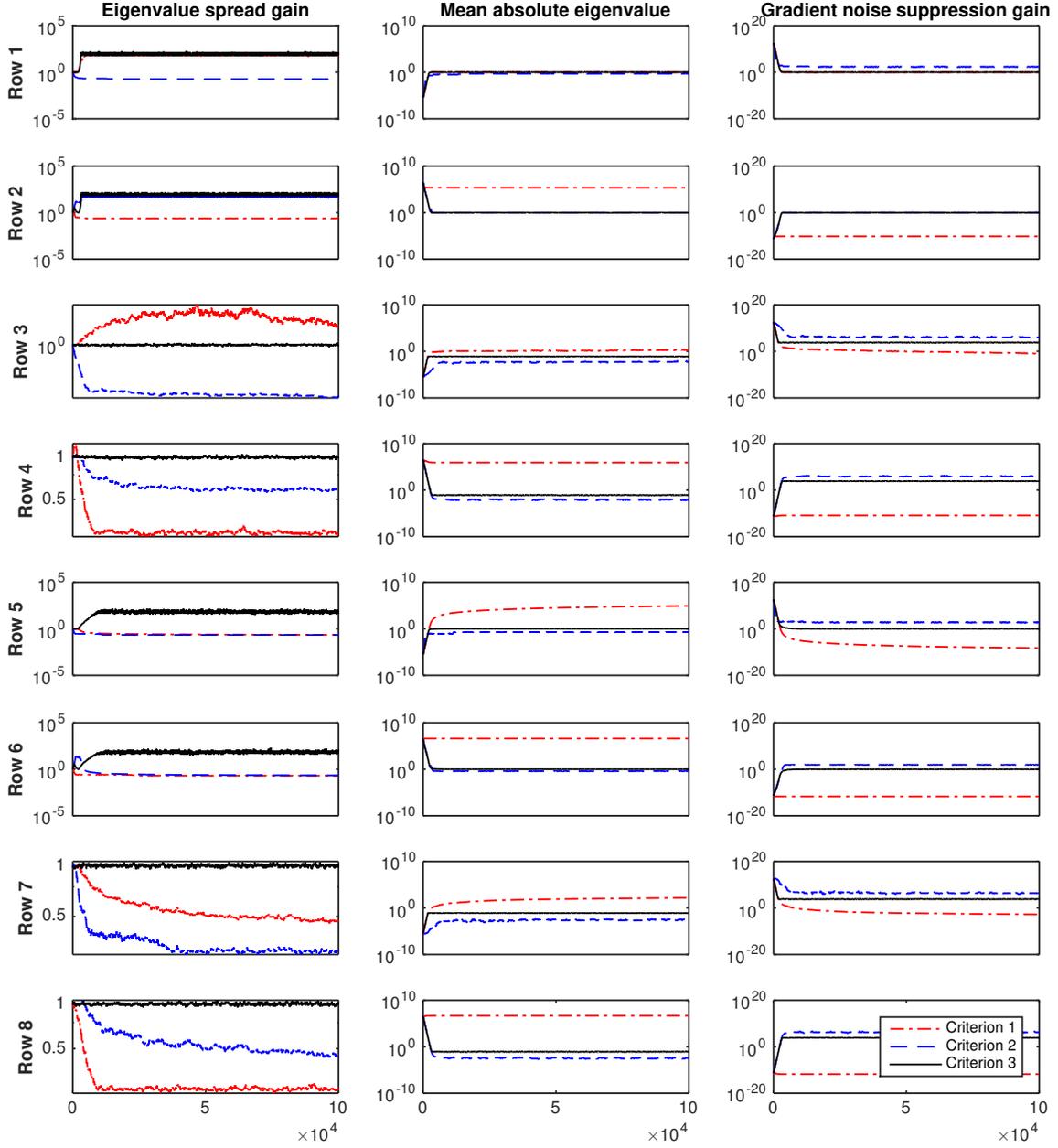}\\
	\caption{ Preconditioner estimation criteria comparison. Row 1: positive definite Hessian, noise free gradient, $\sigma_h^2 = 10^{-12}$; Row 2: positive definite Hessian, noise free gradient, $\sigma_h^2 = 10^{12}$; Row 3: positive definite Hessian, noisy gradient, $\sigma_h^2 = 10^{-12}$; Row 4: positive definite Hessian, noisy gradient, $\sigma_h^2 = 10^{12}$; Row 5: indefinite Hessian, noise free gradient, $\sigma_h^2 = 10^{-12}$; Row 6: indefinite Hessian, noise free gradient, $\sigma_h^2 = 10^{12}$; Row 7: indefinite Hessian, noisy gradient, $\sigma_h^2 = 10^{-12}$; Row 8: indefinite Hessian, noisy gradient, $\sigma_h^2 = 10^{12}$. }
\end{figure*}

\subsection{Blind Equalization (Deconvolution)}

Let us consider a blind equalization (deconvolution) problem using constant modulus criterion \cite{Godard80}. Blind equalization is an important communication signal processing problem, and SGD is a default method. This problem is weakly non-convex in the sense that using a equalizer with enough taps, the constant modulus criterion has no local minimum \cite{Li95}. In our settings, the source signal is uniformly distributed in range $[-1, 1]$, the communication channel is simulated by linear filter $h(z^{-1})=(-0.8+z^{-2})/(1+0.8z^{-2})$, the equalizer, $w(z^{-1})$, is an adaptive finite impulse response (FIR) filter with $21$ taps, initialized by setting its center tap to $1$ and other taps to zeros, and the mini-batch size is $10$. Step sizes of the compared algorithms are manually adjusted such that their steady state intersymbol interference (ISI) performance indices are about $0.027$, where ISI is defined by $\sum_i c_i^2/\max c_i^2-1$, and $\sum_i c_i z^{-1} = h(z^{-1}) w(z^{-1})$. 

Fig.~3 summarizes the results. A preconditioner constructed in the same way as in the proof of Proposition 2 using an adaptively estimated Hessian matrix achieves the fastest convergence, however, such a solution may be too complicated in practice. Precondition 3 performs reasonably well considering its simplicity. Preconditioner 2 completely fails to improve the convergence with step size $0.01$, while a larger step size leads to divergence. Preconditioner 1 does accelerate the convergence due to the weak non-convexity nature of cost function. The plain SGD converges the slowest, and still its steady state ISI is bumpy and higher than $0.027$.   

\begin{figure}[h]
	\centering
	\includegraphics[width=\columnwidth]{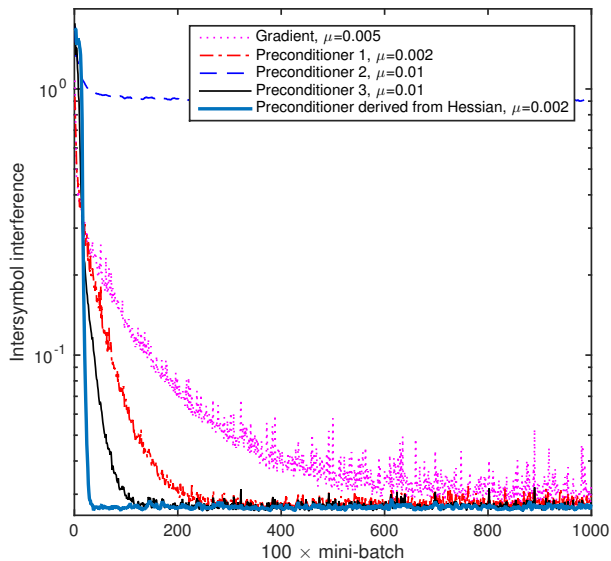}\\
	\caption{  ISI convergence curves. Each point on the curve is an ISI index evaluated every $100$ mini-batches. The curve with preconditioner 2 and step size $0.01$ does not converge, but a larger step size, $0.02$, leads to divergence.  }
\end{figure}

\subsection{A Toy Binary Classification Problem}

In this problem, the input features are $x_1$ and $x_2$, independently and uniformly distributed in $[0, 1]$, and the class label is 
\[ y = {\rm mod} \left [{\rm round}(10^{x_1} - 10^{x_2})/ 2\right], \]
where ${\rm round}(\cdot)$ rounds a real number to its nearest integer, and ${\rm mod}(\cdot)$ denotes modulus after division. Fig.~4 shows the defined zebra stripe like pattern to be learned. It is a challenging classification problem due to its sharp, interleaved, and curved class boundaries. A two layer feedforward neural network with $100$ hidden nodes is trained as the classifier. As a common practice, the input features are normalized before feeding to the network, and the network coefficients of a certain layer are initialized as random numbers with variance proportional to the inverse of the number of nodes connected to this layer. Nonlinear function $\tanh$ is used. Cross entropy loss is the training cost. A $401\times 401$ dense preconditioner is estimated and used in preconditioned SGD. 

Fig.~5 presents one set of typical learning curves with eight different settings. By using the plain SGD as a base line, we find that preconditioner 2 fails to speed up the convergence; both preconditioner 1 and preconditioner 3 significantly accelerate the convergence, and yet preconditioner 3 performs far better than preconditioner 1. Note that there is no trivial remedy to improve the convergence of SGD. A larger step size makes SGD diverge. We have tried RMSProp, and found that it does accelerate the convergence during the initial iterations, but eventually converges to solutions inferior to that of SGD.      

\begin{figure}[h]
	\centering
	\includegraphics[width=\columnwidth]{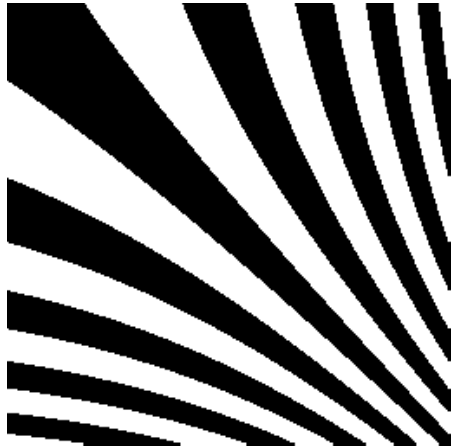}\\
	\caption{The pattern to be learned in the toy example. White areas belong to class 1, and black areas belong to class 0.   }
\end{figure}

\begin{figure}[h]
	\centering
	\includegraphics[width=0.9\columnwidth]{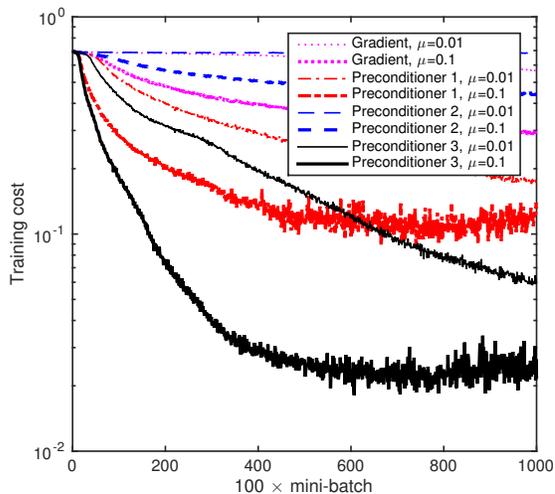}\\
	\caption{ One set of typical convergence curves for the toy binary classification problem.  Each point on the curves is a cross entropy loss averaged over $100$ mini-batches.  }
\end{figure}

\subsection{Recurrent Neural Network Training}

Let us consider a more challenging problem: learning extremely long term dependencies using recurrent neural network. The addition problem initially proposed in \cite{lstm} is considered. The input is a sequence of two rows. The first row contains random numbers independently and uniformly distributed in $[-0.5, 0.5]$. Elements in the second row are zeros, except two of them are marked with $1$. The desired output of the whole sequence is the sum of the two random numbers from the first row marked with $1$ in the second row. More details can be found in \cite{lstm} and our supplemental materials. 

In our settings, the sequence length is $100$, and a standard (vanilla) recurrent neural network with $50$ hidden states are trained to predict the output by minimizing the mean squared error. The feedforward coefficients are initialized as Gaussian random numbers with mean zero and variance $0.01$, and the feedback matrix is initialized as a random orthogonal matrix to encourage long term memories. Backpropagation through time \cite{bptt} is used for gradient calculation.  Totally, this network has $2701$ tunable parameters. A dense preconditioner  might be expensive, and thus a sparse one is used. Coefficients in the first layer naturally form a $50\times 53$ matrix, and coefficients in the second layer form a $1\times 51$ vector. Thus we approximate the preconditioner for gradient of parameters in the first layer as Kronecker product of a $53\times 53$ matrix with a $50\times 50$ matrix, and the preconditioner for gradient of parameters in the second layer is a $51\times 51$ matrix. Preconditioner for gradient of the whole parameter vector is the direct sum of the preconditioner of the first layer and the one of the second layer. 

Fig.~6 summarizes the results of a typical run. Only the preconditioned SGD using preconditioner 3 converges. A recurrent neural network can be regarded as a feedforward one with extremely large depth by unfolding its connections in time. It is known that the issues of vanishing and exploding gradients arise in a deep neural network, and SGD can hardly handle them \cite{Bengio94}. Preconditioned SGD seems perform quite well on such challenging problems. More testing results on the eight pathological recurrent neural network training problems proposed in \cite{lstm} are reported in supplementary materials and \cite{psgd_rnn}, which suggests that preconditioned SGD performs no worse than Hessian-free optimization, although our method has a significantly lower complexity and involves less parameter tweaking than it.   

\begin{figure}[h]
	\centering
	\includegraphics[width=0.9\columnwidth]{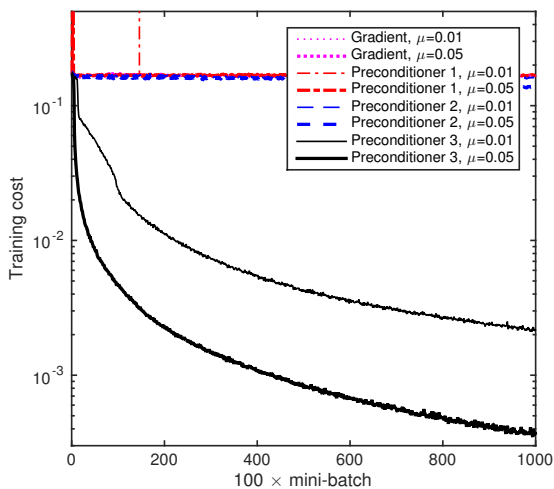}\\
	\caption{ Convergence curves of the addition problem with sequence length $100$. Each point on the curves is a mean squared error averaged over $100$ mini-batches.   }
\end{figure}

\subsection{MNIST Handwritten Recognition}

In the last experiment, we consider the well known MNIST handwritten recognition problem \cite{mnist}. The training data are $60000$ images of handwritten digits, and the test data are $10000$ such images. Feedforward neural network is used as the classifier. The inputs are normalized to dynamic range $[-1,1]$. Due to the high dimension of data, the preconditioners are always approximated as Kronecker products, and each layer has its own preconditioner. 

We first consider a linear classifier using the cross entropy loss. Such a model is also known as a logistic regression model, and it is convex. Fig.~7 summarizes the results. Except that the setting with preconditioner 1 and step size $0.1$ diverges, and the setting with preconditioner 2 and step size $0.01$ converges slow, all the other settings show reasonably good performance. Still, the setting with preconditioner 3 and step size $0.01$ performs the best, converging to a test error rate slightly lower than $0.08$. Here, the test error rate is the ratio of the number of misclassified testing samples to the total number of testing samples.  

\begin{figure}[h]
	\centering
	\includegraphics[width=\columnwidth]{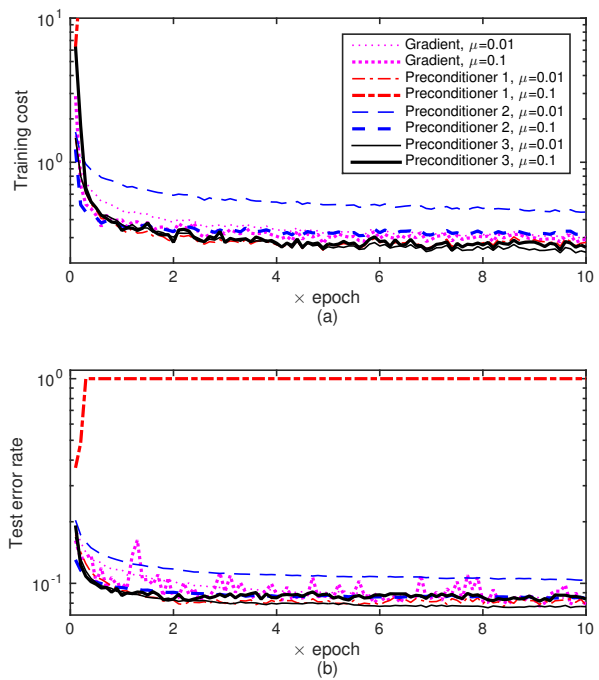}\\
	\caption{ Convergence curves of a linear classifier on MNIST data set.  (a) Each point on the curves is a cross entropy loss averaged over $60$ mini-batches, i.e., $0.1$ epoch. (b) Each point is a test error rate evaluated every $60$ mini-batches. }
\end{figure}

Then we consider a two layer neural network classifier with $300$ hidden nodes, trained using the cross entropy loss as well. Fig.~8 shows the results. Preconditioner 1 and 2 perform poorly due to the non-convexity of cost function. Precondition 3 leads to the best performance on the training data. The best test error rate is slightly higher than $0.02$, achieved by preconditioner 3. However, the neural network trained by preconditioned SGD with preconditioner 3 using a large step size over fits the training data after about two epochs, and the neural network coefficients are pushed to extremely large values. Application related knowledge may be required to prevent or alleviate overfitting. In this example, the input data are rank deficient as many pixels close to image boundaries are zero for all samples. Apparently, this can be one reason causing ill-conditioned Hessian. One common practice is to use regularization terms. We have tried adding regularization term $10^{-4}\pmb \theta^T\pmb \theta$ to the training cost, and found that not only overfitting is avoided, but also the test error rate is reduced to $0.017$ after convergence. Coefficient sharing, as done in convolution and recurrent neural networks, is another way to avoid overfitting. In \cite{psgd_rnn}, a small but deep two dimensional RNN trained with our preconditioned SGD achieves test error rate $0.006$ without using any pre-processing, pre-training, or distorted version of training samples.    

\begin{figure}[h]
	\centering
	\includegraphics[width=\columnwidth]{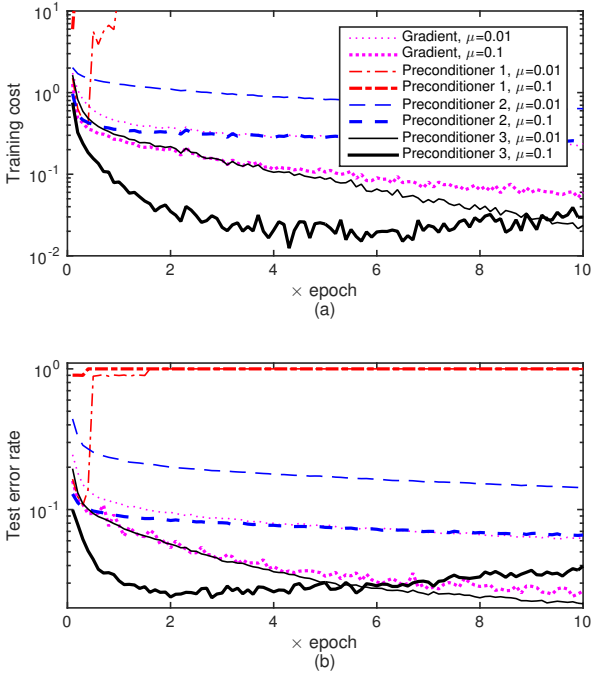}\\
	\caption{ Convergence curves of a two layer neural network classifier on MNIST data set.  (a) Each point on the curves is a cross entropy loss averaged over $60$ mini-batches. (b) Each point is a test error rate evaluated every $60$ mini-batches.  }
\end{figure}

Lastly we train a three layer neural network classifier with $300$ and $100$ nodes in the first and second hidden layers respectively. We deliberately use a multi-class hinge loss to test the numerical robustness of preconditioned SGD in the presence of non-smooth gradient. Supposing the neural network output is ${\pmb o}$ and the target class label is $i$, the hinge loss is defined by $\max(\max_{j\ne i}{o}_j + 1 - {o}_i, 0)$ \cite{mc_hinge_loss}. We use a slightly modified and smoother hinge loss,
\[ \sqrt{ \max(\max_{j\ne i}{o}_j + 1 - {o}_i, 0)^2 + 0.01 } - 0.1.\]
Still, it is not second order differentiable everywhere due to the use of function $\max(\cdot)$. 

Fig.~9 summarizes the results. Again, preconditioner 1 and 2 perform poorly due to the non-convexity of cost function. Precondition 3 considerably accelerates the convergence, and achieves the best test error rate which is slightly higher than $0.02$. With a large step size, the test error rate converges just with about $1.5$ epochs. But unlike the use of cross entropy loss, no apparent overfitting is observed here. Interestingly, the preconditioner estimation algorithms can cope with non-smooth and noisy gradients.    

\begin{figure}[h]
	\centering
	\includegraphics[width=\columnwidth]{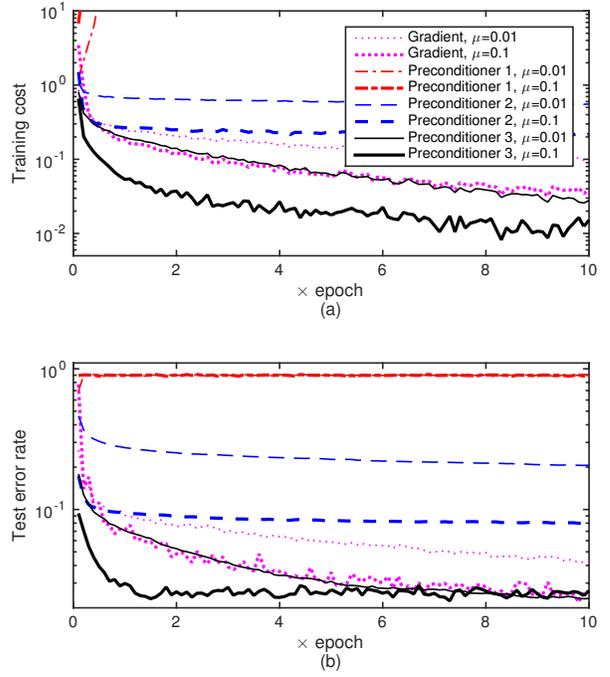}\\
	\caption{ Convergence curves of a three layer neural network classifier on MNIST data set.  (a) Each point on the curves is a multi-class hinge loss averaged over $60$ mini-batches. (b) Each point is a test error rate evaluated every $60$ mini-batches.  }
\end{figure}

We briefly discuss the computational complexity of preconditioned SGD to end this section. Comparison with SGD, preconditioned SGD introduces two fixed overheads per iteration, one more gradient evaluation and one preconditioner update. When simplified preconditioner is used or the gradient calculation is complicated, the overhead due to gradient evaluation dominates. Thus compared with SGD, preconditioned SGD roughly doubles the complexity. Luckily, the two gradients in preconditioned SGD can be evaluated simultaneously to save computational time when parallel computing is available. When the Hessian does not change fast over its parameters, one may evaluate the perturbed gradient and update preconditioner less frequently to save computational complexity as well.   

We give two examples on the time complexity of preconditioner estimation based on profile analysis of the supplementary Matlab code. In the recurrent neural network training example, preconditioner updating only consumes about 0.5\% CPU time. In the MNIST neural network training example, preconditioner updating consumes about 15\% CPU time. In both examples, gradient evaluation dominates the time complexity, and preconditioning can be beneficial as it may accelerate the convergence by far more than two times.  

\section{Conclusions}

Preconditioned stochastic gradient descent (SGD) is studied in this paper. Our analysis suggests that an ideal preconditioner for SGD should be able to reduce the eigenvalue spread and normalize the amplitudes of eigenvalues of the Hessian, and at the same time, avoid amplification of gradient noise. We then study the performance of three preconditioner estimation criteria. Two of them are based on stochastic secant equation fitting as done in the quasi-Newton methods, and naturally they are confined to convex stochastic optimization. A new preconditioner estimation criterion is proposed, and is shown to be applicable to both convex and non-convex optimization problems. We show that the new preconditioner scales the stochastic gradient in a way similar to the Newton method where the inverse of Hessian is the preconditioner, while the other two preconditioners either over or under compensate the gradient due to gradient noise. Based on these criteria, variant stochastic relative (natural) gradient descent preconditioner estimation algorithms are developed. Due to the equivariant property of relative gradient descent, the proposed preconditioner estimation algorithms work well for Hessians with wide numerical dynamic ranges. Finally, both toy and real world problems with different levels of difficulties are examined to study the performance of preconditioned SGD. Using SGD as a base line, we observe that preconditioner 1 may accelerate the convergence for convex and weakly non-convex problems, but inclines to cause divergence; preconditioner 2 seldom causes divergence, but neither can it improve convergence; preconditioner 3 always achieves the best performance, and it provides the only solution capable of training recurrent neural networks requiring extremely long term memories.

\begin{IEEEbiography}
[{\includegraphics[width=1in,height=1.25in,clip,keepaspectratio]{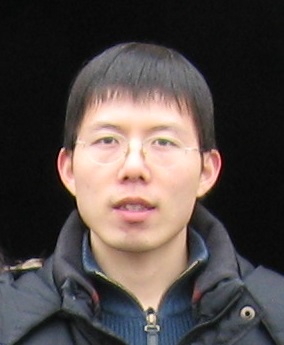}}]
{Xi-Lin Li} received the Bachelor degree and the Master degree, both in electrical engineering, from the Dalian University of Technology, Dalian, China, in 2001 and 2004, respectively, and the Ph.D. degree in control science and engineering from the Tsinghua University, Beijing, China in 2008.
	
From 2008 to 2016, he has worked in the Machine Learning for Signal Processing Lab at the University of Maryland, Fortemedia Inc., and Cisco Systems Inc., successively. His research interests include signal processing, machine learning and optimization.
\end{IEEEbiography}


\begin{thebibliography}{1}

\bibitem{Widrow85}
B.~Widrow and S.~D.~Stearns, \emph{Adaptive Signal Processing}. Englewood Cliffs, New Jersey: Prentice-Hall, Inc., 1985.

\bibitem{Godard80}
D.~Godard, ``Self-recovering equalization carrier tracking in two-dimensional data communications systems,'' \emph{IEEE Trans. Commun.}, vol. 28, no.~11, pp.~1867--1875, Nov. 1980.

\bibitem{Rumelhart86}
D.~E.~Rumelhart, G.~E.~Hinton, and R.~J.~Williams, ``Learning representations by back-propagating errors,'' \emph{Nature}, vol. 323, pp.~533--536, Oct. 1986. 

\bibitem{bptt}
P.~J.~Werbos, ``Backpropagation through time: what it does and how to do it,'' \emph{IEEE Proc.}, vol.~78, no.~10, pp.~1550--1560, Oct. 1990. 

\bibitem{LeCun98}
Y.~LeCun, L.~Bottou, Y.~Bengio, and P.~Haffner, ``Gradient based learning applied to document recognition,'' \emph{Proc. IEEE}, vol.~86, no.~11, pp.~2278--2324, Nov. 1998.

\bibitem{Wray91}
W.~L.~Buntine and A.~S.~Weigend, ``Computing second derivatives in feed-forward networks: a review,'' \emph{IEEE Trans. Neural Netw.}, vol.~5, no.~3, pp.~480--488, May 1991. 

\bibitem{matlab}
H.~Demuth and M.~Beale, \emph{Neural Network Toolbox for Use with MATLAB},  Natick, MA: The MathWorks, Inc., 2002.

\bibitem{Martens2012_hessian_free}
J.~Martens and I.~Sutskever, ``Training deep and recurrent neural networks with Hessian-free optimization,'' In \emph{Neural Networks: Tricks of the Trade}, 2nd ed., vol.~7700, G.~Montavon, G.~B.~Orr, and K.-R.~M\"{u}ller, Ed. Berlin Heidelberg: Springer, 2012, pp.~479--535.

\bibitem{convex_quasi_newton}
N.~N.~Schraudolph, J.~Yu, and S.~G\"{u}nter, ``A stochastic quasi-Newton method for online convex optimization,'' \emph{J. Mach. Learn. Res.}, vol.~2, pp.~436--443, Jan. 2007.

\bibitem{stochastic_newton1}
R.~H.~Byrd, S.~L.~Hansen, J.~Nocedal, and Y.~Singer, ``A stochastic quasi-Newton method for large-scale optimization,'' \emph{SIAM J. Optimiz.}, vol.~26, no.~2, pp.~1008--1031, Jan. 2014. 

\bibitem{stochastic_newton2}
B.~Antoine, B.~Leon, and G.~Patrick, ``SGD-QN: careful quasi-Newton stochastic gradient descent,'' \emph{J. Mach. Learn. Res.}, vol.~10, pp.~1737--1754, Jul. 2009.

\bibitem{rmsprop}
G.~Hinton, \emph{Neural Networks for Machine Learning}. Retrieved from \url{http://www.cs.toronto.edu/~tijmen/csc321/slides/lecture_slides_lec6.pdf}.

\bibitem{Sutskever2013}
I.~Sutskever, J.~Martens, G.~Dahl, and G.~E.~Hinton, ``On the importance of momentum and initialization in deep learning,'' In \emph{30th Int. Conf. Machine Learning}, Atlanta, 2013, pp.~1139--1147.

\bibitem{no_more_pesky_mu}
T.~Schaul, S.~Zhang, and Y.~LeCun, ``No more pesky learning rates,'' arXiv:1206.1106, 2013.

\bibitem{Equilibrated_mu}
Y.~N.~Dauphin, H.~Vries, and Y.~Bengio, ``Equilibrated adaptive learning rates for non-convex optimization,'' in \emph{Advances in Neural Information Processing Systems}, 2015, pp.~1504--1512.

\bibitem{Chunyuan}
C.~Li, C.~Chen, D.~Carlson, and L.~Carin, ``Preconditioned stochastic gradient Langevin dynamics for deep neural networks,'' in \emph{AAAI  Conf. Artificial Intelligence}, 2016.

\bibitem{Carlson}
D.~E.~Carlson, E.~Collins, Y.~P.~Hsieh, L.~Carin, and V.~Cevher, ``Preconditioned spectral descent for deep learning,'' in \emph{Proc. 28th Int. Conf. Neural Information Processing Systems}, Montreal, 2015, pp.~2971--2970. 

\bibitem{hess_kron}
D.~Povey, X.~Zhang, and S.~Khudanpur, ``Parallel training of DNNs with natural gradient and parameter averaging,'' in \emph{Proc. Int. Conf. Learning Representations}, 2015. 

\bibitem{Martens2015}
J.~Martens and R.~B.~Grosse, ``Optimizing neural networks with Kronecker-factored approximate curvature,'' in \emph{Proc. 32nd Int. Conf. Machine Learning}, 2015, pp.~2408--2417.

\bibitem{Cardoso96}
J.-F.~Cardoso and B.~Laheld, ``Equivariant adaptive source separation,'' \emph{IEEE Trans. Signal Process.}, vol.~44, no.~12, pp. 3017--3030, Dec. 1996.

\bibitem{Amari96}
S.~Amari, ``Natural gradient works efficiently in learning,'' \emph{Neural Computation}, vol.~10, no.~2, pp.~251--276, Feb. 1998.

\bibitem{Li95}
Y.~Li and Z.~Ding, ``Convergence analysis of finite length blind
adaptive equalizers,'' \emph{IEEE Trans. Signal Process.}, vol.~43, no.~9, pp.~2120--2129, Sept. 1995.

\bibitem{lstm}
S.~Hochreiter and J.~Schmidhuber, ``Long short-term memory,'' \emph{Neural Computation}, vol.~9, no.8, pp.~1735--1780, 1997.

\bibitem{Bengio94}
Y.~Bengio, P.~Simard, and P.~Frasconi, ``Learning
long-term dependencies with gradient descent is
difficult,'' \emph{IEEE Trans. Neural Netw.}, vol.~5, no.~2, pp.~157--166, Mar. 1994.

\bibitem{psgd_rnn}
X.-L. Li, ``Recurrent neural network training with preconditioned stochastic gradient descent,'' arXiv:1606.04449, 2016.

\bibitem{mnist}
Y.~LeCun, C.~Cortes, and C.~J.~C.~Burges, \emph{THE MNIST DATABASE}. Retrieved from \url{http://yann.lecun.com/exdb/mnist/}.

\bibitem{mc_hinge_loss}
K.~Crammer and Y.~Singer, ``On the algorithmic implementation of multiclass kernel-based vector machines,'' \emph{J. Mach. Learn. Res.}, vol.~2, pp.~265--292, 2001.

\end{thebibliography}
\end{document}